# Automatic Rail Component Detection Based on AttnConv-Net

Tiange Wang, Zijun Zhang, *Senior Member, IEEE*, Fangfang Yang, and Kwok-Leung Tsui

*Abstract*—The automatic detection of major rail components using railway images is beneficial to ensure the rail transport safety. In this paper, we propose an attention-powered deep convolutional network (AttnConv-net) to detect multiple rail components including the rail, clips, and bolts. The proposed method consists of a deep convolutional neural network (DCNN) as the backbone, cascading attention blocks (CAB), and two feed forward networks (FFN). Two types of positional embedding are applied to enrich information in latent features extracted from the backbone. Based on processed latent features, the CAB aims to learn the local context of rail components including their categories and component boundaries. Final categories and bounding boxes are generated via two FFN implemented in parallel. To enhance the detection of small components, various data augmentation methods are employed in training process. The effectiveness of the proposed AttnConv-net is validated with one real dataset and another synthesized dataset. Compared with classic convolutional neural network based methods, our proposed method simplifies the detection pipeline by eliminating the need of prior- and post-processing, which offers a new speed-quality solution to enable faster and more accurate image-based rail component detections.

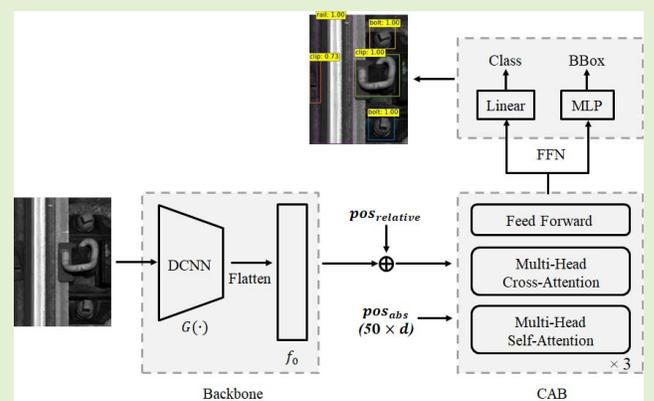

## I. INTRODUCTION

AUTOMATIC inspection of railway system (AIRS) is a crucial task in railway maintenance. Any anomalies that happened to the railway system may potentially lead to train derailments and other accidents. Various tasks related to AIRS have been addressed in previous studies, such as the detection of rail track irregularity [1]–[3], essential components [4], and obstacles [5], [6]. Early inspection relies on the experienced maintenance crews walking along rails to check the health condition of rails manually and periodically [7], which is time-consuming and laborious. Due to the development of sensing technologies, AIRS starts to be explored based on a variety of collected signals including the vibration [5], GPS [1], and acoustic emissions [8]–[10]. However, accuracies of signal-based methods are limited by the quality of signal collection and the surrounding environment. Certain sensors, such as stereo cameras and LIDAR, are prohibitively expensive and require considerable power in applying them into AIRS.

The recent advancement of image processing techniques has driven the development of vision-based approaches for AIRS, directly affecting the inspection accuracy and speed. The earlier type of vision-based AIRS benefits from the emerging development of machine learning (ML) principles. It is capable of processing raw images captured by visual sensors and converting them into handcrafted features, such as the Haar-like features [7], Harris-Stephen and Shi-Tomasi features [11], Gabor filter algorithm and its variants [12], [13], etc. Another classic classifier, the support vector machine (SVM), is among the most frequently utilized ML algorithms in AIRS. In [14], a two-layer SVM network was proposed to detect damages on rail surfaces. The first layer was used to extract the rail area and the second layer was used as a classifier to distinguish categories of surface defects on the rail. The combination of SVM and different feature extractions was also widely applied to address AIRS [15], [16].

So far, most of the ML-based AIRS solutions focus on explicitly engineered features, especially texture analysis. It is

Manuscript received November 2, 2021; revised November 29, 2021; accepted November 29, 2021. Date of publication December 3, 2021; date of current version January 31, 2022. This work was supported in part by the Hong Kong Research Grants Council Theme-Based Research Scheme Project under Grant T32-101/15-R, in part by the Research Impact Fund Project under Grant R5020-18, and in part by the General Research Fund Project under Grant 11215418, and in part by the CityU Strategic Research under Grant 7005302. The associate editor coordinating the review of this article and approving it for publication was Prof. Yu-Dong Zhang. *(Corresponding author: Zijun Zhang.)*

Tiange Wang and Zijun Zhang are with the School of Data Science, City University of Hong Kong, Hong Kong (e-mail: zijzhang@cityu.edu.hk).

Fangfang Yang is with the School of Intelligent Systems Engineering, Sun Yat-sen University, Guangzhou 510275, China.

Kwok-Leung Tsui is with the Grado Department of Industrial and Systems Engineering, Virginia Polytechnic Institute and State University, Blacksburg, VA 24061 USA.

Digital Object Identifier 10.1109/JSEN.2021.3132460

challenging to deal with more complex cases with different illuminations or flexible component sizes.

The rich accumulation of ML-based approaches and the rapid development of deep learning (DL) technologies results in the evolution of AIRS to overcome the requirement of higher efficiency on detection tasks [17]. DL-based approaches form the mainstream in most existing studies regarding specific detection targets, such as defects, components, and obstacles [18]. Rail surface defects mainly include surface cracks and rolling contact fatigue wears, which are usually small and irregularly distributed. By cropping the defect regions out as a pre-processing step, sizes of the defects were fixed and the distribution of various defects was not as dense as before [2], [19], [20]. Rail fasteners are essential components, preventing rail overturning and longitudinal movement. The visual inspection of fasteners is primarily for detecting the absence of or damage to fasteners. Similar to the defect detection, some of approaches rely on pre-processing of the region proposals. For example, by defining the regions of interest (ROI), other components including the anchor and tie were located in the proposed areas [21]. ROI was also combined with SVM to classify tie plates [22]. These DL-based methods mainly served the single class detection, requiring the uniform size and similar targets. Even like the multi-class detection task in [21], it failed to predict different types of objects simultaneously. Moreover, additional inputs, such as the distance and ROI, were required to implement the inspection. The obstacle detection of railways also benefits from DL-based methods. In [23], a fusion refine neural network (FR-Net) was presented to detect obstacles ahead in the railway shunting mode. FR-Net applied a two-step strategy that a coarse detection module first refined locations and prior anchors, and then a finer detection module was applied to achieve more accurate object locations and classifications. A similar strategy was presented in [24] that a differential feature fusion convolutional neural network (DFF-Net) was introduced to detect traffic obstacles on the railway tracks in the shunting mode. It was composed of two modules, the prior object detection module for generating initial anchor boxes and the differential feature fusion sub-module for enriching the semantic information. In [25], a dynamic programming method was presented to extract the train course and frontal railroad track. By using the video data, rails on both sides were identified according to the vanishing point. In [26], the laser scanner and stereo cameras were applied to develop a 3D obstacle detection system. Considering the insufficient training samples, recent obstacle detections focused on reconstructing normal data and subsequently identified anomalies with relatively high reconstruction errors [27]–[31]. However, they tend to reconstruct the input image into a lower-quality version without explicitly imposing constraints on the authenticity of the reconstructions.

In general, we notice that the existing methods on AIRS still rely on feature extractions through DCNNs. Some approaches even require additional inputs, such as the distance, laser, or sonic data. Prior- and post-processing steps are also needed to generate a sufficient number of candidate boxes and remove redundant boxes, making the model tuning process complicated.

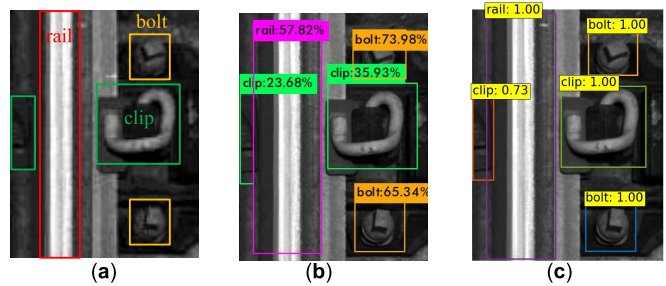

Fig. 1. **(a)** Detection targets in considered AIRS including rail, clip, and bolt. Rails are always located vertically in the image, clips and bolts are symmetrically distributed on both sides of the rail. **(b)** Visualized detection results with traditional DL-based method. **(c)** Detection via AttnConv-net results in much higher confidence scores on clips and bolts, as well as a more appropriate bbox to rail.

To realize multi-class detection more concisely without any pre- and post-processing, we develop an attention-powered deep convolutional network (AttnConv-net) consisting of a deep CNN structure (DCNN) as a backbone, cascading attention blocks (CAB), and two feed forward networks (FFN). It simplifies the computational process and enhances the detection accuracy of rail components, regarding much variety on the component size. Although the attention mechanism [32] has achieved a considerable success in natural language processing, its advantages in addressing AIRS are untapped. In addition to its scarcity in railway applications, the attention mechanism for detection tasks always follow a standard architecture of the same depths of encoder and decoder block [33], [34]. We challenge this convention by using CAB instead of a complete encoder-decoder architecture to output categories and locations (in terms of bounding boxes). This work explores the advantages of adopting CAB on rail component detection task. It is capable of building correlations between elements in feature maps that can be treated as the sequence of attention inputs. Moreover, the characteristics of CAB determine that the scale of ultimate detections can be appropriately decided due to different purposes, saving the cost on introducing prior conditions and simplifying the computation cost.

The detection targets in the paper are displayed in Fig. 1. Usually, a rail is located vertically and crosses the whole image. Clips and bolts are symmetrically distributed on both sides of the rail with smaller sizes. The left clip in Fig. 1 (a) is even truncated. In this case, we are committed to improving the overall detection efficiency as well as focusing on strengthening the detection of small and truncated components. As visualized in Fig. 1 (c), the detector we propose reaches higher confidence scores for all components than the conventional DL-based method, indicating that the proposed AttnConv-net can still achieve a satisfactory detection performance even if the threshold is raised.

The main contributions of this work are summarized as follows:

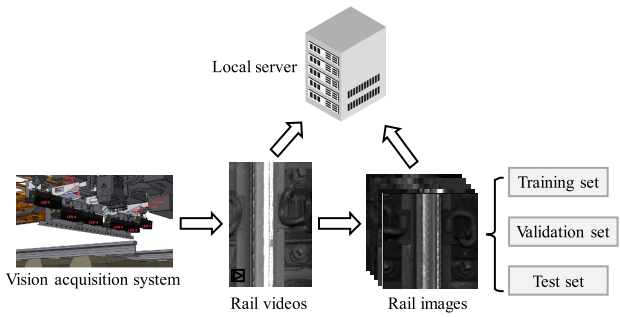

Fig. 2. The data collection process of $D_I$.

1. A novel AttnConv-net is proposed to enable the automatic image-based rail component detection and enhance its accuracy while maintaining a low computational complexity. Through ablation studies, we demonstrate that the proposed CAB in AttnConv-net is capable of effectively distinguishing different components via learning local context. The simple architecture of CAB leads to less computational consumption and faster detection speed.
2. We analyze the benefit of data augmentation on detecting truncated and small components. By enriching the diversity of training rails with additional augmentation methods, the detection of clips, which is a more challenging task, can be significantly enhanced.
3. Based on the collected dataset, our method significantly improves 13.0% and 13.4% accuracy against benchmarks on detecting small clips and bolts, respectively. By incorporating the CAB with DCNN, the model yields the highest accuracy of 61.9% with an average speed of 32.2ms. A synthesized dataset is also employed to validate the applicability of the proposed AttnConv-net on the studied AIRS problem. The computational results on both datasets outperform a set of state-of-the-art benchmarking methods.

## II. THE RAIL SURFACE IMAGE DATASET

The rail surface image dataset $D_I$ was provided by an industrial partner in Hong Kong. Fig. 2 shows the data collection process. The high speed cameras in a real-time track monitoring system [4] continuously take videos of the rail surface at a diagnostic run of the train with a low speed, 15 km/h. By splitting the videos into frames, we obtain the rail surface images and store them in the local server. Due to changes in the background, the brightness and size of the rail images vary a lot. Fig. 1(a) displays the target components, including the rail, clips, and bolts. The rail is the largest component, which is located vertically and owns the height of the whole image. Clips and bolts are symmetrically distributed on both sides to fasten the rails. Due to vibrations on devices, the clip can be partially captured with a relatively small ground-truth box and an edge location while the bolts can be entirely included in the images because of smaller sizes. As showed in Fig. 2, the collected dataset $D_I$ is divided into three sets, in which 691 images are utilized for training, 345 images for validation, and 350 for testing. Total 6451 components from three categories are annotated with ground-truth bounding boxes including the category, center coordinates, width, and height. Basically, each image includes at least one rail track with a varying number of bolts and clips. A rail image contains at most 9 components in our dataset.

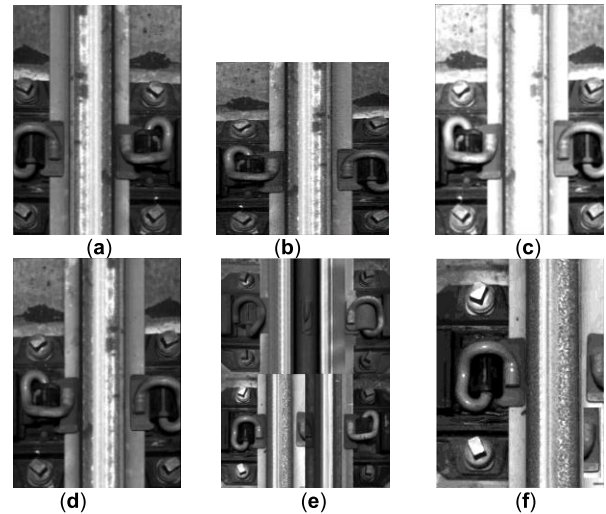

Fig. 3. Two groups of data augmentation methods. One refers basic photometric distortion and geometric distortion: **(a)** Mirror flip. **(b)** Rescaling to 640 × 640. **(c)** Exposure change (brighter). **(d)** Saturation change. Another refers methods for improving the detection of small component: **(e)** Stitcher method, four different images of similar size are stitched in spatial dimension. **(f)** Copy-pasting method, bolts and truncated clips are copied and pasted twice without overlap with any existing component.

Compared with generic multi-class object detection problems, targets in the rail images usually do not overlap, which makes the AIRS problem more specific. Nevertheless, there are still two challenges in the rail component detection. First, the collected dataset is relatively small compared to public domain dataset, such as COCO and PASCAL VOC. Thus, available samples for training are limited. Secondly, a considerable part of clips in the dataset is truncated. It is challenging to detect truncated components due to their significantly smaller sizes and more edge positions. Therefore, we employed data augmentation methods to overcome the limitation of $D_I$ as well as enhance the detection performance on small components. Data augmentation is conducted via two groups of methods. One group refers to the basic photometric distortion and geometric distortion [35], such as flipping, rescaling, exposure change, and saturation change. Another group of augmentation methods are applied to improve the small component detection, including the stitcher [36] and copy-pasting [37]. Fig. 3 displays samples of data augmentation via these two groups of methods. Fig. 3 (a-d) includes the mirror flip, rescaling, exposure change, and saturation change. Fig. 3 (e) is a sample of the stitcher method. Four different rail images are stitched in spatial dimension as one training sample. The width and height of each image are scaled to 1/2 of the averaged width and height of four images. By stitching them together, four different contexts are mixed and learned at the same time. Moreover, the composite image is about the same size as the initial one. Clips that were originally at the edge are located in the middle of the image with significantly smaller normalized size, which is potential to strengthen the detection of edge-side and small components. Fig. 3 (f) shows a sample of copy-pasting. The number of truncated clips and bolts is increased by pasting them to different positions of the same image so that the contribution of small objects into computing the loss function is improved. Before pasting, the copied component is not scaled. Each bolt or truncated clip is pasted twice. The principle of pasting aims to prevent any overlaps with existing components. It is possible that some bolts and truncated clips in the same image might not be tripled due to the limited space. After data augmentation, the number of training targets is expanded to twice of the collected volume.

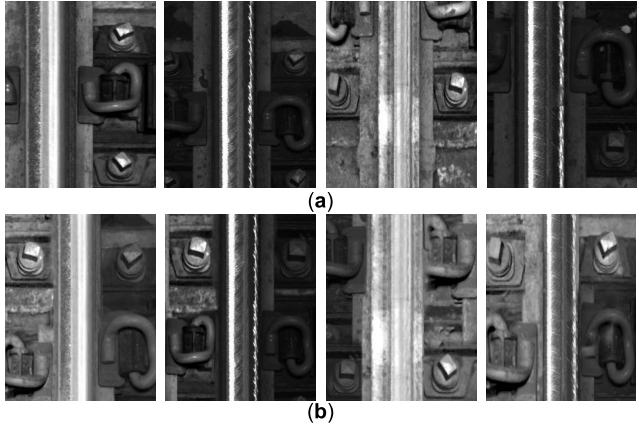

Fig. 4. Examples of synthesized images in $D_{II}$. **(a)** The collected rail images in $D_I$. **(b)** By randomly combing different components, the illumination in each synthesized image varies in different areas.

TABLE I
A SUMMARY OF DATASETS UTILIZED IN EXPERIMENTS

| Datasets | Sets | # images | # components (w/o data augmentation) | | | # components (w/ data augmentation) | | |
|---|---|---|---|---|---|---|---|---|
| | | | Rail | Bot | Clip | Rail | Bolt | Clip |
| $D_I$ | training | 691 | 694 | 1240 | 1263 | 694 | 2095 | 2250 |
| | validation | 345 | 347 | 638 | 595 | 347 | 1303 | 1179 |
| | test | 350 | 356 | 706 | 612 | 356 | 1360 | 1227 |
| $D_{II}$ | test | 500 | 503 | 1149 | 1214 | | / | |

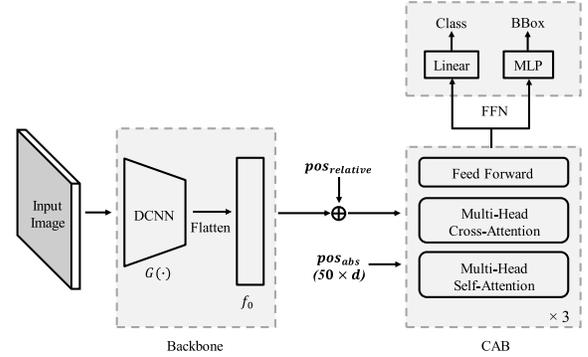

Fig. 5. The overall architecture of AttnConv-net.

To further validate the applicability of the proposed AttnConv-net on AIRS, a new dataset $D_{II}$ containing a set of synthesized rail images is applied in comparative analyses via computational studies. Based on the collected dataset $D_I$, we randomly combine the components in different rail images to form a new image. The combination of different contexts results in a more complex background as well as a more challenging dataset. Fig. 4 displays examples of the synthesized images. There are 500 images and 2866 components contained in $D_{II}$. They are utilized to evaluate advantages of the AttnConv-net with benchmarking methods in Section IV.C. The two datasets utilized in our experiments are summarized in Table I.

## III. METHOD DESCRIPTION

The overall architecture of the AttnConv-net is depicted in Fig. 5. It is mainly composed of three parts: 1) a DCNN backbone for extracting a compact feature representation, 2) CAB for selectively discriminating the DCNN output, and 3) two parallel FFN for predicting categories and locations of different rail components, respectively. The developed AttnConv-net aims to achieve a high accuracy and low model complexity for AIRS problem simultaneously.

The whole framework is illustrated as follows. The input of AttnConv-net is an RGB rail image $x \in \mathbb{R}^{3 \times H_0 \times W_0}$. Let the function $G(\cdot)$ denotes the DCNN architecture, lower-resolution feature maps $f \in \mathbb{R}^{C \times H \times W}$ are obtained via $f = G(x)$. Our experiments mainly present the ResNet-50 as the backbone, which is one of the classic DCNN architectures, so that $H = \frac{H_0}{32}, W = \frac{W_0}{32}, c = 2048$ according to [38]. Then, a $1 \times 1$ convolution is applied to reduce the channel dimension of the high-level activation map $f$ from $c$ to a smaller dimension $d = 512$. Spatial dimensions are collapsed into one dimension, resulting in a two-dimensional feature map $f_0 \in \mathbb{R}^{HW \times d}$. The sequence of attention input $f_0$ is defined as follows:

$$f_{out}(i, j, d_m) = f_{in}(i, j, c) * F_m(1, 1, c) \quad (1)$$
$$f_0[H * W, d] \triangleq f_{out}[H, W, d] \quad (2)$$

where $F_m$ denotes the $m$-th filter weight with depth $c$, and $d_m$ denotes the $m$-th channel of the output feature map.

In the proposed framework, $f_0$ with fixed positional embedding $pos_{relative}$ directly passes through the CAB, in which each element in the sequence learns an alignment to gather from others. After the concatenation with $pos_{relative}$, the CAB takes learned positional embedding $pos_{abs}$ as additional inputs. To avoid autoregressive, the number of input tokens of CAB is determined by the length of $pos_{abs}$ for realizing the parallel computation. The two types of positional embeddings are defined as follows:

$$pos_{relative}^{(i,j)} = \begin{cases} \sin\left(\dfrac{i}{10000^{j/d}}\right), & j = 2t \\ \cos\left(\dfrac{i}{10000^{(j-1)/d}}\right), & j = 2t+1 \end{cases} \quad (3)$$

$$pos_{abs} \in \mathbb{R}^{50 \times 512} \sim N(0, 0.1^2) \quad (4)$$

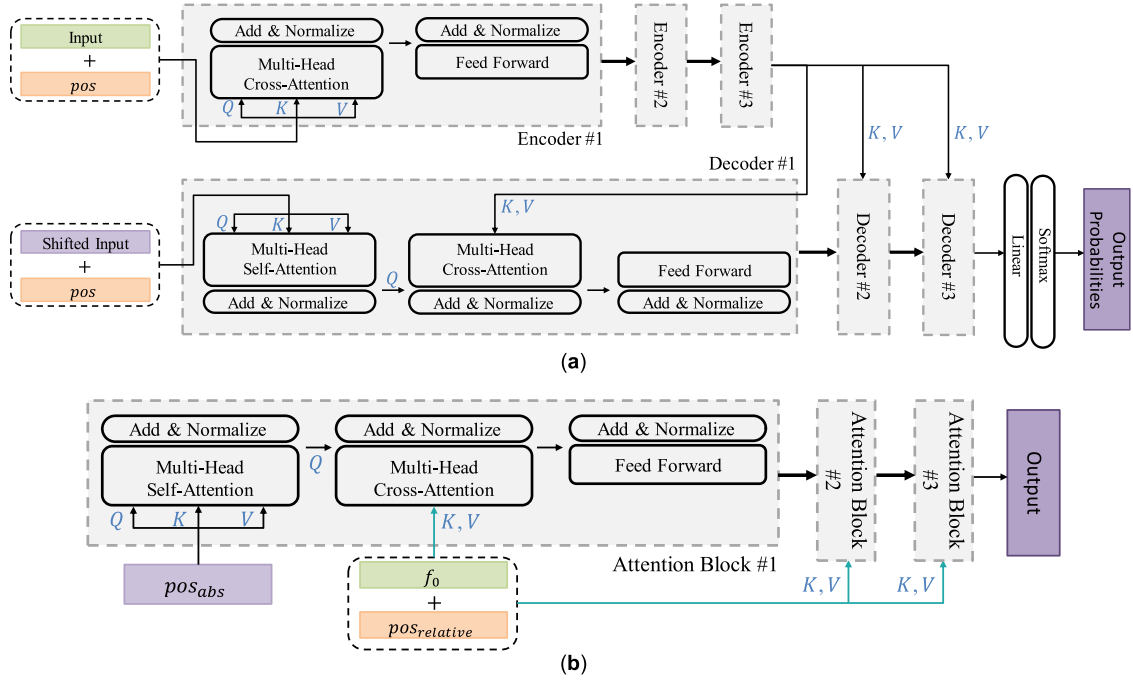

Fig. 6. Illustrating the differences between **(a)** existing encoder-decoder architecture [32] and **(b)** the proposed CAB architecture. A standard encoder-decoder architecture consists of the same depths of encoder and decoder. However, in our method, the CAB reduces the computational cost by directly employing $pos_{abs}$ instead of shifted input information to learn the spatial relation of rail components. Besides, the linear and softmax layers are removed at the end of CAB.

where $i \in \{0, 1, \ldots, HW-1\}$, $j \in \{0, 1, \ldots, d-1\}$. Since the CAB is permutation-invariant, information indicating the spatial positions must be defined before the detection. In $pos_{relative}$, sinusoidal positional embedding is applied to inject relative positional information into vectors. The $pos_{abs}$ is randomly initialized following a normal distribution $N(0, 0.1^2)$. With $pos_{relative}$ and $pos_{abs}$, the position features on components with the positional embedding are continuously added to each layer in CAB.

Fig. 6 (b) presents the detailed architecture of CAB. Each attention block contains a multi-head self-attention layer $A_{self}(\cdot)$ for passing $pos_{abs}$, a multi-head cross-attention layer $A_{cross}(\cdot)$ for building the information transfer with backbone, and a feed forward layer for projecting the refined matrix to a larger space and extracting the required information more easily. The matrix calculation of $A_{self}(\cdot)$ generates three matrices from the same input, the queries ($Q$), keys ($K$), and values ($V$). A single-head self-attention layer is computed as follows:

$$Z^i_{single} = A^i_{self}(X_{input}) = softmax(\frac{Q^i(K^i)^T}{\sqrt{d_k}})V^i \quad (5)$$

where $Q^i = X_{input}W^i_Q$, $K^i = X_{input}W^i_K$, and $V^i = X_{input}W^i_V$. Note that $W^i_Q \in \mathbb{R}^{d \times d_k}$, $W^i_K \in \mathbb{R}^{d \times d_k}$, and $W^i_V \in \mathbb{R}^{d \times d_v}$. In the self-attention layer, $Q$, $K$, and $V$ perform the computation in parallel and yield a $d_v$-dimensional output $Z^i_{single}$. Multiple $Z^i_{single}$ are concatenated and output a final $Z_{multi}$ with another weight $W_O \in \mathbb{R}^{hd_v \times d}$. We follow [32] and set $h = 8$, $d_k = d_v = d/h = 64$. $A_{cross}(\cdot)$ works in a similar way. The only difference is that $Q$ is passed from the last self-attention layer while $K$ and $V$ are passed from the backbone, which plays a role in information transfer. In the developed AttnConv-net, the first $A_{self}(\cdot)$ layer in the first attention block takes only $pos_{abs}$ as the input for generating $Q$, which will be fed into the next cross-attention layer. In each $A_{cross}(\cdot)$ layer, $K$ and $V$ are calculated using different weights. Compared with existing encoder-decoder architecture [32] as shown in Fig. 6 (a), the proposed CAB employs the $pos_{abs}$ instead of the shifted input information to learn the spatial relation of rail components. Besides, the linear and softmax layers are removed at the end of attention blocks, resulting an output that can be processed in the next FFN module. The proposed CAB has fewer parameters, which greatly reduces the computational consumption and boosts the speed of detection. With the last output from CAB, the categories and bounding boxes of rail components are detected through a linear projection and a multi-layer perceptron (MLP) executed in parallel, respectively.

The loss function of AttnConv-net follows the common detection methods, including a linear combination of $-\log[\hat{p}(c)]$ for category predictions and $L_{box}(b, \hat{b})$ for the similarity between predicted bounding boxes and ground-truths. Specifically, $L_{box}(b, \hat{b})$ is composed of the L1 loss of the center coordinates and $giou(b, \hat{b})$ of the normalized sizes. Since the FFN module outputs categories and bounding boxes in parallel, a pair-wise matching cost is defined to find a bipartite matching between the output $\hat{p}(c)$ and $\hat{b}$ within index $\hat{\sigma}(i)$ and ground-truth $y_i$ with the lowest cost:

$$L_{match}(y, \hat{y}) = \sum_{i=1}^{50}\left[-\mathbb{1}_{\{c_i \neq \emptyset\}}\hat{p}_{\hat{\sigma}(i)}(c_i) + \mathbb{1}_{\{c_i \neq \emptyset\}}L_{box}(b_i, \hat{b}_{\hat{\sigma}(i)})\right] \quad (6)$$

$$\hat{\sigma} = \arg\min(L_{match}(y, \hat{y})) \quad (7)$$

$$L(y, \hat{y}) = \sum_{i=1}^{50} \left[ -\log \hat{p}_{\hat{\sigma}(i)}(c_i) + 1_{\{c_i \neq \emptyset\}} L_{box}(b_i, \hat{b}_{\hat{\sigma}(i)}) \right] \quad (8)$$

where $\hat{p}_{\hat{\sigma}(i)}(c_i)$ is the probability of category $c_i$ and $\hat{b}_{\hat{\sigma}(i)}$ is the predicted bounding box. Since the number of predictions was set larger than the maximal number of ground-truths, there are always some predictions indicating 'no object'. Finding a valid permutation of predictions that can pair with limited ground-truths thus becomes important. This procedure plays the same role as the heuristic assignment rules used to match region proposals [39] or prior anchors [40] to ground-truth targets in generic detectors. By finding one-to-one matching for the prediction set without duplicates, the post-processing step such as NMS is no longer needed. With $\hat{\sigma}$, the ultimate loss is derived in (8).

The overall training procedure is displayed in **Algorithm 1**. Rail images as well as ground-truth labels including category $C$ and bounding box $B$ are utilized for training the AttnConv-net. The sigmoid function $mlp(\cdot).sigmoid$ and the linear projection $linear(\cdot)$ are employed to predict bounding boxes and categories, respectively. $L(y, \hat{y})$ is derived for every batch of inputs. If the loss value gets infinite, the training procedure stops automatically. Otherwise, the gradient $\nabla \theta_t$ is calculated via $bp(\cdot)$ and model weights $\theta_t$ are updated with a learning rate $\eta$.

---

**Algorithm 1** Training of the AttnConv-Net

**Input**: Training set $X$, training labels $\{C, B\}$.
**Output**: Model parameter $\theta$
1. **for** epoch **in** range(epochs) **do**
2.   **for** j **in** range(training_volume // batch_size) **do**
3.     $f = G(X_j)$;
4.     $f_0 = f * F(1, 1, c)$;
5.     $h = CAB(pos_{relative}(f_0) + f_0, pos_{abs})$;
6.     $\hat{P}(C) = linear(h)$;
7.     $\hat{B} = mlp(h).sigmoid$;
8.     $L(y, \hat{y}) = \sum \left[ -\log \hat{p}(c_i) + L_1(b_i, \hat{b}_i) + giou(b_i, \hat{b}_i) \right]$;
9.     **if** $L(y, \hat{y}) = \infty$ **then**
10.       **break**;
11.     **else**
12.       $\nabla \theta_t = bp(loss = L(y, \hat{y}))$;
13.       $\theta_{t+1} = \theta_t - \eta \nabla \theta_t$;
14.     **end if**
15.   **end for**
16. **end for**
17. **return** $\theta_{t+1}$;

---

## IV. COMPUTATIONAL EXPERIMENTS

In this section, we first report the training setup of experiments. Then, evaluation metrics exploited to assess the detection capability of AttnConv-net are described in detail. Finally, results and analysis via ablation studies as well as comparisons with a set of benchmarking methods are provided.

### A. Training Setup

We mainly apply ResNet-50 as the DCNN backbone in the computational experiments. Three attention blocks make up the CAB module. The AttnConv-net is trained using the AdamW optimizer with momentum of 0.9 and weight decay of 0.0005. The initial learning rate of the backbone and CAB are both fine-tuned to 0.0001. A linear scaling rule is applied for the learning rate to drop at 250 epochs by a factor of 10. Since the batch normalization weights in backbone are frozen when using pretrained network structure, a dropout rate set to 0.1 is only adopted to layers in CAB and FFN parts. Weights of AttnConv-net are initialized with a normalized initialization. We set the training schedule to 300 epochs on three Nvidia Geforce RTX2080 Ti GPUs. Each GPU is responsible for processing 2 images at one time so that the batch size should be 6. It takes nearly 4 hours to finish the training of AttnConv-net.

### B. Evaluation Metrics

The primary evaluation metric in this work is the average precision (AP) over 10 intersection over union (IoU) thresholds that are evenly distributed between 0.5 and 0.95. AP is defined as follows:

$$\text{AP} = \frac{1}{10} \sum_{iou \in \{0.5, 0.55, \ldots, 0.95\}} \text{AP}^{(\text{IoU})} \quad (9)$$

$$\text{AP}^{(\text{IoU})} = \sum (r_{n+1}^{(\text{IoU})} - r_n^{(\text{IoU})}) p_{interp}(r_{n+1}^{(\text{IoU})}) \quad (10)$$

$$p_{interp}(r_{n+1}^{(\text{IoU})}) = \max_{\tilde{r} \geq r_{n+1}^{(iou)}} p(\tilde{r}) \quad (11)$$

where $p(\tilde{r})$ represents the relation between precision and recall. In addition to using AP to represent the accuracy of locating and classifying targeted components, APs under IoU = 0.50 (AP@50) and IoU = 0.75 (AP@75) are also compared to clarify the detection capacity. The precision and recall required for $p(\tilde{r})$ are defined as follows:

$$precision = \frac{TP}{TP + FP} \quad (12)$$

$$recall = \frac{TP}{TP + FN} \quad (13)$$

where *TP* denotes the true positives, *FP* denotes the false positives, and *FN* denotes the false negatives. Since the number of predictions is fixed, the precision and recall values are discrete. $\text{AP}^{(\text{IoU})}$ is easy to obtain by measuring the exact area under the precision-recall curve. The metrics above are all averaged across different categories, in which we primarily focus on small objects, including clips and bolts. If not specified, AP from the last training epoch is reported as the validation AP. Besides, we use the average detection time on the test set to compare the processing speed of the AttnConv-net on each image. Billion floating point operations (BFLOPs) that are accumulated over multiple calculations and a number of parameters are also recorded to estimate the algorithm complexity.

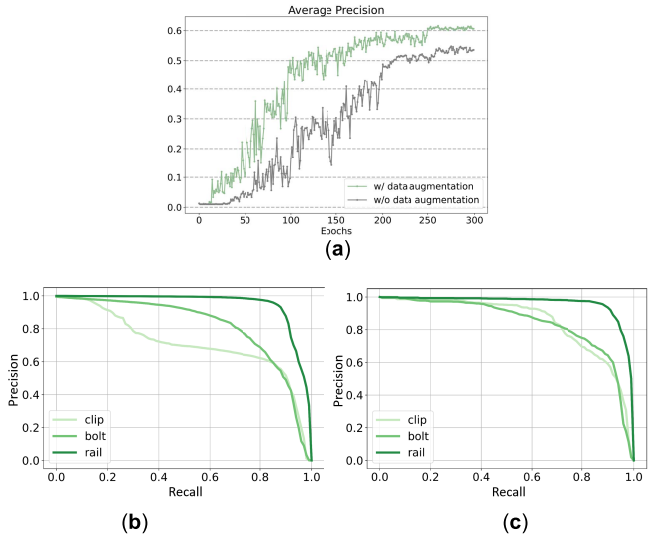

Fig. 7. **(a)** The learning curves of validation AP with and without data augmentation. **(b)** Precision-recall curves with basic data augmentation methods. **(c)** Precision-recall curves with extra data augmentation methods, including stitcher and copy-pasting. The detection perfomance on clips has significantly improved, with no damages on the detection of rails and bolts.

TABLE II
VALIDATION RESULTS OF DIFFERENT PREDICTION SIZES

| # Predictions | # Params | AP | AP@75 | AP@50 |
|---|---|---|---|---|
| 10 | 28.65M | 57.5 | 63.1 | 90.7 |
| 20 | 28.66M | 58.7 | 65.3 | 92.3 |
| 50 | 28.66M | **61.3** | 68.5 | **95.9** |
| 80 | 28.67M | 61.2 | **68.9** | 95.6 |
| 100 | 28.68M | 61.1 | 68.0 | 95.2 |

## C. Results and Analysis

*1) Performance Optimization:* Since the training volume of the rail dataset $D_I$ is doubled with data augmentation technologies, the effect of data augmentation is first explored. Fig. 7 displays the impact of data augmentation methods on the validation set. The overall detection results with and without data augmentation are compared in Fig. 7 (a). AP value over multiple thresholds has significantly increased with various augmentation methods. Since two groups of augmentation methods are employed, we further compare their effect on three different categories. As shown in Fig. 7 (b-c), extra data augmentation methods including the stitcher and copy-pasting have positive impacts on the detection of clips while has no damages on the detection of rails and bolts.

Besides the impact of data augmentation, the influence of prediction size is also evaluated in Table II. Since the learned positional embedding $pos_{abs}$ is the initial input of the CAB, the prediction size should be the same as the sequence length of $pos_{abs}$, which is decided by characteristics of the attention layers. Although the maximum number of objects in a rail image of our dataset is 9, Table II shows that setting a large number of predictions with more slack (such as 50) can effectively increase the validation AP by almost 4% and improves the AP@50 by 5.2 % compared with 10 predictions. When the prediction size increases, the AttnConv-net is capable of maintaining the same accuracy but accompanying more computations. Therefore, we set 50 as the optimal prediction size in the remaining experiments.

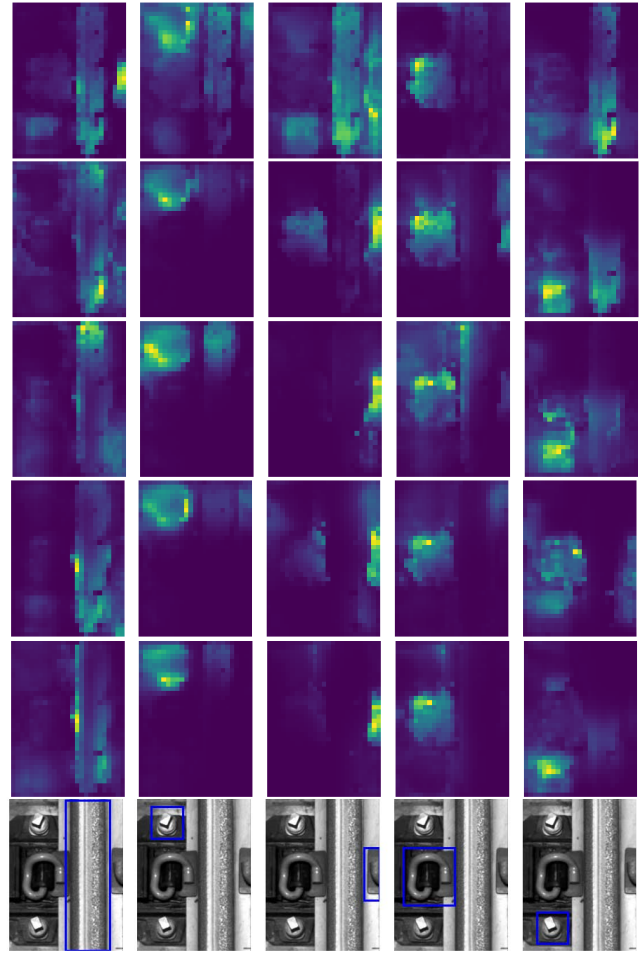

Fig. 8. Attention weights of the last block in CAB. From top to bottom, the depths of CAB are $n = 1, 2, 3, 6, 9$, respectively. The last row shows the ground-truth bounding boxes. As the attention level deepens, the CAB focuses on specific components rather than on the entire image.

Finally, the attention weights from the last block of CAB are visualized in Fig. 8. From top to bottom, the attention weights belong to the CAB with different number of attention blocks $n = 1, 2, 3, 6, 9$, respectively. The last row provides the test image and labels of different components. For each component, the CAB puts the attention on the relative part for predicting specific bounding box and category. As the attention level deepens, the CAB focuses on specific components rather than on the entire image. The visualization of attention weights gives the intuition that using $n = 3$ could attain the same attention effect as using deeper attention blocks in the proposed AttnConv-net. By directly concatenating the backbone and CAB, a new speed-quality baseline is provided for the future AIRS toward a faster and more accurate detection. The remaining experiments also prove that the simplified configuration attains comparable detection performances as adding additional encoder blocks.

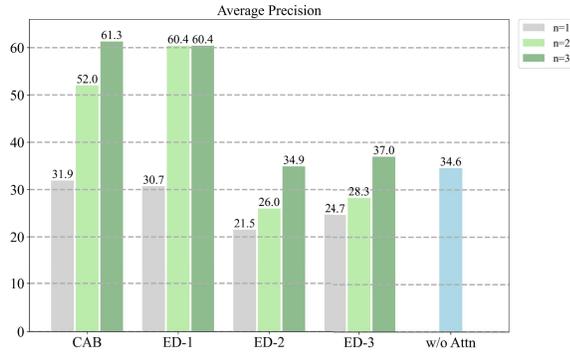

Fig. 9. Validation AP under different AttnCov-net variants. When $n = 3$, the CAB yields the highest AP with the least parameters and complexity, while the ED-1 variant achieves comparable performance but with more computations.

TABLE III
EFFECT OF EACH COMPONENT ON APs

| Backbone | CAB | FFN | AP | AP@75 | AP@50 |
|---|---|---|---|---|---|
| ✓ | ✓ |   | 47.5 | 56.2 | 83.0 |
| ✓ |   | ✓ | 34.6 | 46.9 | 79.3 |
| ✓ | ✓ | ✓ | **61.3** | **68.5** | **95.9** |

*2) Ablation Studies:* We analyze the effectiveness of each part of AttnConv-net. The experiments are conducted on $D_I$. By default, the experiment settings and the module structure remain the same as Section IV. C 1).

*a) Effect of CAB:* Since the first introduction of the attention mechanism, many prior works have assumed a standard attention architecture in detection tasks, which has the same number of encoder and decoder blocks [32]–[34]. Therefore, we develop two AttnCov-net variants as follows. (1) The ED variant replaces the CAB with standard encoder-decoder architecture as in DETR [33], where the initial input is the sum of $f_0$ and $pos_{relative}$. The $pos_{abs}$ is treated as the input of decoder part. (2) The w/o Attn variant removes CAB from AttnConv-net, where the features $f_0$ from backbone is directly fed to the FFN module. We evaluate the effectiveness of CAB by changing the number of attention block $n$. The experiment results are shown in Fig. 9. ED-1, ED-2, and ED-3 represent the variants with 1, 2, and 3 encoder blocks, respectively. We find that CAB with $n = 3$ achieves the best performance among all variants of AttnCov-net. The comparison between CAB and w/o Attn variant justifies the advantage of using the attention mechanism to learn local context. Moreover, the comparison between CAB and ED shows that the proposed CAB is capable of effectively detecting different components with less parameters and computation burden. Although ED has additional encoder blocks, it learns from the same input as CAB, which is considered as redundant learning in the AttnConv-net. Additionally, with more encoder blocks, the detection accuracy of the ED variants drops sharply with comparable $n$, revealing that advantages of the encoder blocks are overestimated in the AIRS. Experiments with larger $n$ are also conducted. Their results on AP performance are comparable to $n = 3$.

*b) Effect of each part in AttnConv-net:* To evaluate the effectiveness of each part in AttnConv-net, we design two AttnConv-net variants as follows. (1) The w/o Attn variant, which is the same as the variant mentioned before. (2) The w/o FNN variant removes the FFN module completely, leaving only the CAB to implement detections. Since the backbone is essential for the feature extraction and cannot be removed, we compare both two variants with the full AttnConv-net model. Results are presented in Table III. By comparing the w/o Attn with the full AttnConv-net, we can observe that the proposed CAB accounts for most of the AP values, losing almost half AP without CAB. It proves that the attention based mechanism significantly contributes to the detection of rail components. By comparing w/o FNN with full AttnConv-net, the better performance of full AttnConv-net model proves the validity of adding FFN to predicting categories and locations of components.

*3) Comparative Experiments:* A comprehensive comparison between the AttnConv-net and benchmarking methods based on $D_I$ and $D_{II}$ is demonstrated in Table IV. YOLO series [2] and RetinaNet [40] are trained for a longer schedule of 10k iterations with a batch size of 64 to achieve optimal detection performances. DETR models [33] are trained for 500 epochs with a batch size of 6. The training of each model on three GPUs takes at least 12 hours, which is triple the time consumed by our proposal. Moreover, the data augmentation strategies mentioned in Section II are applied to all benchmarking methods. Backbones utilized in the comparative experiments, including Vgg, ResNet, and Darknet, are all pre-trained on ImageNet [35]. Table IV shows that the proposed AttnConv-net generally owns the highest accuracies on both $D_I$ and $D_{II}$. It achieves at most 13.6% higher AP value than the optimized benchmarks, with a comparable inference time of 32.2ms. Especially for detecting small objects, it owns at most 13.0% higher AP in detecting clips and 13.4% higher in detecting bolts. The experiment results imply that the attention mechanism in the CAB enhances the capability of AttnConv-net for learning the category, location, and size of various components based on both collected dataset $D_I$ and synthesized dataset $D_{II}$. Moreover, as the CAB can effectively localize and categorize rail components with simple attention blocks, the memory complexity of the AttnConv-net is greatly reduced. The fast speed and accurate performance of AttnConv-net make it paramount for the widespread adoption in not only the railway inspection but also potentially in more real-world applications.

## V. CONCLUSION

In this work, a novel attention-powered deep convolutional network for component detection in AIRS was developed to simplify the detection pipeline as well as reduce the computations. It operated effectively to attain more advanced detection performance on different rail datasets. By applying data augmentation methods, the AttnConv-net achieved 61.9 AP and 60.2 AP on the test sets of $D_I$ and $D_{II}$, respectively. In detecting small or truncated clips and bolts, the CAB helped capture local contexts to better localize and categorize them. The proposed AttnConv-net made a lot of efforts on reducing the computational complexity and memory capacity. Its detection speed was comparable to the DETR and YOLO methods and was significantly faster than RetinaNet. The experiments revealed the limitation of full encoder-decoder architectures in detection task, and proved that the simple CAB was capable of realizing impressive detection performances well as reducing the model complexity. With the detection framework simplified, we offered a new speed-quality solution, realizing faster and more accurate rail component detection.

TABLE IV
COMPARISON WITH BENCHMARKS ON TEST SETS

| Model | Backbone | BFLOPs/Time (ms) | # Params | $D_I$-test | | | | $D_{II}$-test | | | |
|---|---|---|---|---|---|---|---|---|---|---|---|
| | | | | AP | Rail | Clip | Bolt | AP | Rail | Clip | Bolt |
| YOLOv2 | darknet19 | 70/23.8 | 35.1M | 50.1 | 59.8 | 46.3 | 44.2 | 48.0 | 57.9 | 43.7 | 42.8 |
| YOLOv3 | darknet53 | 155/34.5 | 58.3M | 49.2 | 53.6 | 47.2 | 46.8 | 48.4 | 53.2 | 46.6 | 45.6 |
| DETR | resnet50 | 69/33.9 | 30.4M | 60.6 | 69.1 | 57.0 | 55.9 | 59.8 | 69.4 | 55.9 | 53.8 |
| DETR | vgg16 | 116/67.0 | 46.2M | 57.0 | 62.7 | 54.2 | 55.1 | 55.7 | 62.3 | 52.8 | 52.4 |
| RetinaNet | resnet101 | 208/58.8 | 83.5M | 48.3 | 53.5 | 45.9 | 45.5 | 47.2 | 53.0 | 45.8 | 42.9 |
| AttnConv-net | resnet50 | 65/32.2 | 28.7M | **61.9** | **69.3** | 58.9 | **57.6** | 59.4 | **70.4** | 54.2 | 53.8 |
| AttnConv-net | darknet53 | 101/37.6 | 29.7M | 61.6 | 69.1 | **59.1** | 56.3 | **60.2** | 70.0 | **55.9** | **54.6** |

Besides the successful detection for rail components, the potential of AttnConv-net on benefiting other rail track condition monitoring applications via analyzing railway images needs to be further explored. We will focus on small surface defects of AIRS in future work, which can also benefit from the attention mechanism. Another challenge for the proposed method is further decreasing the model complexity and speeding up the detection, such that the real-time inspection can be implemented with videos as input.

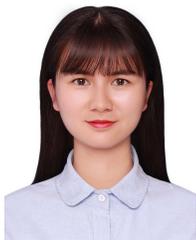

**Tiange Wang** received the B.Sc. degree in industrial engineering from Nanjing University, Nanjing, China, in 2018. She is currently pursuing the Ph.D. degree with the School of Data Science, City University of Hong Kong, Hong Kong, SAR, China. Her research interests include railway health monitoring, intelligent transportation, and machine learning.

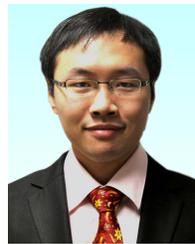

**Zijun Zhang** (Senior Member, IEEE) received the B.Eng. degree in systems engineering and engineering management from The Chinese University of Hong Kong, Hong Kong, SAR, China, in 2008, and the M.S. and Ph.D. degrees in industrial engineering from The University of Iowa, Iowa City, IA, USA, in 2009 and 2012, respectively.

Currently, he is an Associate Professor with the School of Data Science, City University of Hong Kong. His research interests include machine learning and computational intelligence with applications in renewable energy, facility energy management, and intelligent transportation domains. He serves as an Associate Editor for IEEE TRANSACTIONS ON SUSTAINABLE ENERGY, IEEE POWER ENGINEERING LETTERS, and *Journal of Intelligent Manufacturing*.

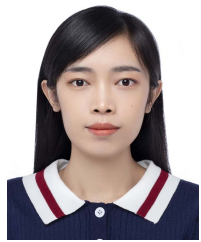

**Fangfang Yang** received the Ph.D. degree in systems engineering and engineering management from the City University of Hong Kong in 2017. She is an Associate Professor at the School of Intelligent Systems Engineering, Sun Yat-sen University. Her research interests include prognostics and health management, degradation modeling, remaining useful life prediction, and deep learning.

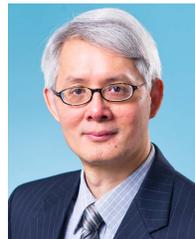

**Kwok-Leung Tsui** received the Ph.D. degree in statistics from the University of Wisconsin–Madison, Madison, WI, USA, in 1986.

He was the Chair Professor of Industrial Engineering with the School of Data Science and the Department of Systems Engineering and Engineering Management, City University of Hong Kong, Hong Kong, from 2009 to 2020; a Professor/an Associate Professor with the School of Industrial and Systems Engineering, Georgia Institute of Technology, Atlanta, GA, USA, from 1990 to 2011; and a member of Technical Staff at the Quality Assurance Center, AT&T Bell Labs, Holmdel, NJ, USA, from 1986 to 1990. He is currently a Professor with the Grado Department of Industrial and Systems Engineering, Virginia Polytechnic and State University, Blacksburg, VA, USA. His current research interests include data science and data analytics, surveillance in healthcare and public health, personalized health monitoring, prognostics and systems health management, calibration and validation of computer models, process control and monitoring, and robust design and Taguchi methods.

Dr. Tsui is a Fellow of the American Statistical Association, the American Society for Quality, the International Society of Engineering Asset Management, and the Hong Kong Institution of Engineers; an elected Council Member of the International Statistical Institute, and a U.S. Representative to the ISO Technical Committee on Statistical Methods.